\documentclass{article} % For LaTeX2e
\usepackage{iclr2021_conference,times}

\usepackage{amsmath,amsfonts,bm}

\def\eqref#1{equation~\ref{#1}}
\def\1{\bm{1}}

\DeclareMathAlphabet{\mathsfit}{\encodingdefault}{\sfdefault}{m}{sl}
\SetMathAlphabet{\mathsfit}{bold}{\encodingdefault}{\sfdefault}{bx}{n}

\newcommand{\Cov}{\mathrm{Cov}}
\usepackage{hyperref}
\usepackage{url}

\title{PCE-PINNs:\\Physics-Informed Neural Networks for\\Uncertainty Propagation in Ocean Modeling}

\author{Bj{\"o}rn L{\"u}tjens\thanks{Corresponding email: \texttt{lutjens@mit.edu}}  $^{\;,1}$, Catherine H. Crawford$^2$, Mark Veillette$^3$, Dava Newman$^1$  \\
$^1$Human Systems Laboratory, MIT ,
$^2$IBM Research, 
$^3$MIT Lincoln Laboratory
}

\usepackage{graphicx} % more modern
\usepackage{amsfonts}
\usepackage{amsmath,soul}
\usepackage{amssymb}
\usepackage[capitalize]{cleveref} %cref
\usepackage{dsfont} % mathds
\usepackage[font=small,skip=0pt]{subcaption} %subfigure
\usepackage{balance}
\usepackage[font=small]{caption}
\usepackage{bm} % bold italic vectors
\usepackage{wrapfig}
\usepackage{float} % To place figure on top right as float

\usepackage{color}
\def\name/{\textit{PCE-PINNs}}

\usepackage[authormarkuptext=name,addedmarkup=bf,authormarkupposition=left]{changes}
\definechangesauthor[name={J.~H.}, color={blue}]{jh}
\definechangesauthor[name={M.~E.}, color={red}]{me}
\definechangesauthor[name={B.~L.}, color={green}]{bl}
\usepackage{tikz,mathtools}
\crefformat{equation}{(#2#1#3)}
\Crefformat{equation}{Equation~(#2#1#3)}
\Crefname{equation}{Equation}{Equations}

\usetikzlibrary{shapes,positioning,automata,arrows,fit,backgrounds,calc}
\tikzstyle{block} = [draw, fill=blue!20, rectangle,minimum height=1em,
minimum width=2em] \tikzstyle{sum} = [draw, fill=blue!20, circle, node
distance=1cm] \tikzstyle{input} = [coordinate] \tikzstyle{output} =
[coordinate] \tikzstyle{pinstyle} = [pin edge={to-,thin,black}]
\usetikzlibrary{trees} \usetikzlibrary{decorations.pathmorphing}
\usetikzlibrary{decorations.markings}
\definecolor{darkgreen}{rgb}{0,0.5,0}
\definecolor{darkred}{rgb}{220,20,60}

\newcommand{\cmmnt}[1]{\ignorespaces}

\newcommand{\bit}{\begin{itemize}}
\newcommand{\ei}{\end{itemize}}

\iclrfinalcopy % Uncomment for camera-ready version, but NOT for submission.
\begin{document}

\maketitle

\begin{abstract}
Climate models project an uncertainty range of possible warming scenarios from $1.5$ to $5^\circ$ global temperature increase until 2100, according to the CMIP6 model ensemble. Climate risk management and infrastructure adaptation requires the accurate quantification of the uncertainties at the local level. 
Ensembles of high-resolution climate models could accurately quantify the uncertainties, but most physics-based climate models are computationally too expensive to run as ensemble.
Recent works in physics-informed neural networks (PINNs) have combined deep learning and the physical sciences to learn up to 15k faster copies of climate submodels. However, the application of PINNs in climate modeling has so far been mostly limited to deterministic models. 
We leverage a novel method that combines polynomial chaos expansion (PCE), a classic technique for uncertainty propagation, with PINNs. The PCE-PINNs learn a fast surrogate model that is demonstrated for uncertainty propagation of known parameter uncertainties. We showcase the effectiveness in ocean modeling by using the local advection-diffusion equation.
%Given new measurements, the resulting surrogate model can quickly interpolate the measurements. 
\end{abstract}
%!TEX root=main.tex

\section{Introduction}\label{sec:introduction}
Informing decision-makers about uncertainties in local climate impacts requires ensemble models. Ensemble models solve the climate model for a distribution of parameters and initial conditions to generate a distribution of local climate impacts~\citep{Gneiting_2005}. The physics of most oceanic processes can be well modeled at high-resolutions, but generating large ensembles is computationally too expensive: High-resolution ocean models resolve the ocean at $8-25km$ horizontal resolution and require multiple hours or days per run on a supercomputer~\citep{Fuhrer_2018}. 
Recent works are leveraging physics-informed deep learning to build ``surrogate models``, i.e., computationally-lightweight models that interpolate expensive simulations of ocean, climate, or weather models~\citep{Rasp_2018,Brenowitz_2020,Yuval_2020b,Kurth_2018, Runge_2019}. The lightweight models achieve a accelerate the simulations on the order of $30-15k$-times~\citep{Yuval_2020b, Rackauckas_2020}. Building lightweight surrogate models could enable the computation of large ensembles. 

The incorporation of domain knowledge from the physical sciences into deep learning has recently achieved significant success~\citep{Raissi_2019,Brunton_2019,Rasp_2018}.%, Lutjens_2021, Cachay_2021}. 
Within physics-informed deep learning one could adapt the neural network architecture to incorporate physics as: inputs~\citep{Reichstein_2019}, training loss~\citep{Raissi_2019}, the learned representation~\citep{Lusch_2018, Greydanus_2019, Bau_2020}, hard output constraints~\citep{Mohan_2020}, or evaluation function~\citep{Lutjens_2021, Lesort_2019}. Alternatively, one could embed the neural network in differential equations~\citep{Rackauckas_2020}, for example, as: parameters~\citep{Garcia_2006, Raissi_2019}, dynamics~\citep{Chen_2018}, residual~\citep{Karpatne_2017, Yuval_2021}, differential operator~\citep{Raissi_2018, Long_2019}, or solution~\citep{Raissi_2019}. We embed a neural network architecture in the  solution which will enable fast propagation of parameter uncertainties and sensitivity analysis. 

While most work in physics-informed deep learning has focused on deterministic methods, recent methods explore the expansion to stochastic differential equations~\citep{Zhang_2019,Dandekar_2021,Yang_2020a,Zhu_2019,Yang_2020b}. In particular,~\citet{Zhang_2019} achieves lightweight surrogate models for parameter estimation and uncertainty propagation by combining physics-informed neural networks~\citep{Raissi_2019} with arbitrary polynomial chaos~\citep{Wan_2006}. %We relax the assumption of necessary parameter observations, 
We use the simpler polynomial chaos expansion~\citep{Smith_2013} instead of arbitrary polynomial chaos expansion, and focus on the task of uncertainty propagation in~\citep{Zhang_2019}.
Further, we are the first in applying the combination of polynomial chaos and neural networks to the stochastic local advection-diffusion equation (ADE). Methods of uncertainty quantification have extensively been demonstrated on the local ADE~\citep{Smith_2013}; the advantage of neural networks is the ability to estimate PCE coefficients in high-dimensional spaces. The local ADE, also called horizontally averaged Boussinesq equation, is more challenging than the 1D stochastic diffusion equation from~\citep{Zhang_2019} and illustrates the application to ocean modeling.

In summary our work contributes, \name/, the first method for fast uncertainty propagation of parameter uncertainties with physics-informed neural networks in ocean modeling.
%!TEX root=main.tex

\section{Approach}\label{sec:approach}
We are defining the initial value problem of solving the stochastic partial differential equation,
\begin{align*}
\mathcal{N}_z [T(t,z;\omega);\kappa(z;\omega)] &= 0,\; t,z\in D, \omega\in\Omega, \\
\text{B.C.: } \mathcal B_{t,z}[T(t=0,z;\omega)] &= 0, \\
%\mathcal B_{t,z}[T(t,z;\omega)] &= 0, z\in \Gamma,\\
%\mathcal B_{t,z}[\frac{T(z,t;\omega)}{dz}] &= 0, z\in \Gamma,\\
\end{align*}
with spatial domain, $D_z$, temporal domain, $D_t$, random space, $\Omega$, domain boundary, $\Gamma$, nonlinear operator, $\mathcal{N}$, and Dirichlet boundary conditions, $\mathcal{B}_{t,z}$.

\subsection{Defining the local advection-diffusion equation}
\begin{figure}[t]%tbp!]%t]
 \centering
 \begin{subfigure}{.4\columnwidth}
  \centering
      \includegraphics [trim=0 0 0 0, clip, width=1.\textwidth, angle = 0]{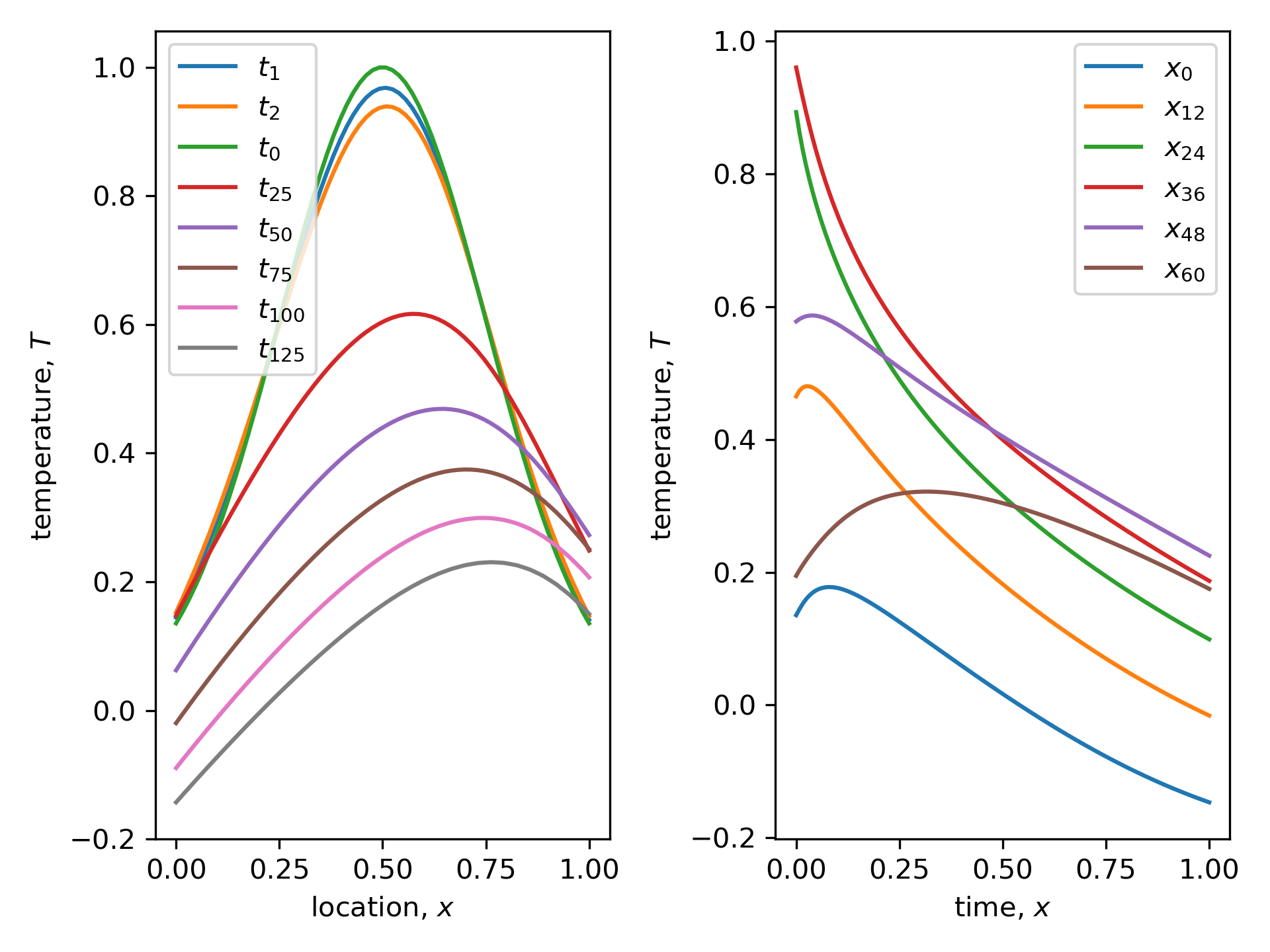}
    %\caption{Target solution}
  \end{subfigure}
\caption[Solution]{\textbf{Local advection diffusion equation.} The local advection diffusion equation simulates how the initial temperature profile, $T(t=0, z)$ (green), is distributed over time. We can observe that the positive diffusivity, $\kappa>0$, flattens the temperature curve (left) and the vertical velocity, $w>0$, shifts the curve to the right over time.} \label{fig:sample_solution} 
\vspace{-.25in}
\end{figure}
We are given the local advection-diffusion equation which models the temperature distribution in a vertical ocean column over time, 

\begin{equation}
f = \frac{\delta T(t,z;\omega)}{\delta t} + \frac{\delta}{\delta z}(wT(t,z;\omega)) - \frac{\delta}{\delta z}\left(\kappa(z;\omega) \frac{\delta T(t,z;\omega)}{\delta z}\right)
\label{eq:ade}\end{equation}
with height, $z\in D_z=[0,1]$, time, $t \in D_t=[0,1]$, source, $f=0$, noise, $\omega \in \Omega$, temperature, $T:D_t,D_z \rightarrow \mathds R$, stochastic diffusivity, $\kappa(z,\omega)$, constant vertical velocity, $w=10$. 

We assume that the distribution over the diffusivity is known, for example, through data assimilation or Bayesian parameter estimation. Specifically, the diffusivity is assumed to follow an exponential Gaussian process (GP) with $\kappa(z;\omega)=\exp({Y_\kappa}(z;\omega))$. The GP, $Y_\kappa(z; \omega)$, is defined by mean, $\mu_{Y_\kappa} = 1000$, correlation length, $L=0.3$, variance, $\sigma_{Y_\kappa} = 1.0$, exponent, $p_\text{GP}=1.0$, and a covariance kernel that is similar to the non-smooth Ornstein-Uhlenbeck kernel: 

\begin{equation}
\begin{aligned}
  \text{Cov}_{Y_\kappa}(z_1, z_2) = \sigma_{Y_\kappa}^2 \exp(-\frac{1}{p_{GP}} (\frac{\lvert z_1 - z_2 \rvert}{L})^p).\\
\end{aligned}  
\label{eq:kernel}
\end{equation}

\subsection{Polynomial chaos expansion in neural networks}
In practice, computing ensembles of differential equations such as in equation~\cref{eq:ade} for a distribution of parameter inputs is often computationally prohibitive. Hence, we aim to learn a copy, or fast surrogate model, of the differential equation solver, $\hat T: D_x \times D_z \rightarrow \mathds R$, assuming a known parameter distribution and a set of ground truth solutions, $T \in \mathds T$, from the solver. 

The polynomial chaos expansion (PCE) approximates arbitrary stochastic functions by a linear combination of polynomials~\citep{Smith_2013}. The polynomials capture the stochasticity by applying a nonlinear function to typically simple distributions and the coefficients capture the spatio-temporal dependencies~\citep{Smith_2013}. PCE has been widely adopted in computational fluid dynamics (CfD) community, because it offers fast inference time, analytical statistical summaries, such as $C_0=\mu_T$, and the theoretical guarantees of polynomials~\citep{Smith_2013}. However, the computation of PCE coefficients,  $C_{\vec\alpha}(t, z)$, is analytically complex, because the computation differs among problems, and computationally expensive, because the computation involves integrals over the random space~\citep{Smith_2013}. %, as described in detail in [\blXX{appendix}]. 
Hence, we leverage neural networks to learn the PCE coefficients, $\hat C_{\vec\alpha}(t, z)$, directly from observations of the solution. 

The polynomial chaos expansion then approximates the solution as,
\begin{equation}
\begin{aligned}
\hat T(t,z;\vec\xi) = \sum_{j=0}^{\lvert A \rvert} \hat C_{\vec\alpha_j}(t,z) \Psi_{\vec\alpha_j}(\xi_1, ..., \xi_n)
\end{aligned}  
\label{eq:poly_basis}
\end{equation}
with the NN-based PCE coefficients, $\hat C_{\vec\alpha}(t, z)\in\mathds R$, the vector of polynomial degrees or multi-index, $\vec\alpha_j\in A$ with $j\in\{0, ..., \lvert A\rvert\}$, the set of multi-indices, $A$, the maximum polynomial degree, $n$, and the set of polynomials, $\Psi_{\vec\alpha}(\vec\xi)$. The polynomials are defined by a set of multivariate orthogonal Gaussian-Hermite polynomials, 
\begin{equation}
\begin{aligned}
\Psi_{\vec \alpha_j}(\xi_1, ..., \xi_n) &= \Pi_{i=1}^n \psi_{\alpha_{ji}}(\xi_i),\\
&= \Pi_{i=1}^n \text{He}_{\alpha_{ji}}(\xi_i), \\
\text{with } \xi_i &\sim \mathcal{N}(0,1).
\end{aligned}
\label{eq:gauss_herm_poly}
\end{equation}
with the one-term (monic) polynomials, $\psi_{\alpha_{ji}}$ of polynomial degree, $\alpha_{ji}$. We are choosing the random vector of each stochastic dimension, $i\in\{0, ..., n\}$, to be a Gaussian, $\xi_i \sim \mathcal N(0,1)$ and use the associated probabilists' Hermite polynomials, $\text{He}_{\alpha_{ji}}$. The polynomial degrees are given by the total-degree multi-index set, $A=\{\vec \alpha_j \in \mathds{N}_0^{n}: \lvert \lvert \vec \alpha_j \rvert\rvert_1 = \sum_{i=1}^n\alpha_{ji} \leq n\}$. For example, $A = \{[0,0], [0,1], [1,0], [1,1], [2,0], [0,2]\}$ for $n=2$. The number of terms, $\lvert A\rvert$, is given by, $\lvert A\rvert= \frac{(n+n)!}{n!n!}$.

The neural network (NN) then jointly approximates all PCE coefficients, 
\begin{equation}
\begin{aligned}
NN_{C_{A}}(t, z):=\hat C_{A}(t, z):\mathds D_t \times D_z \rightarrow \mathds{R}^{\lvert A \rvert}.
\end{aligned}
\label{eq:nn_pce_coefs}
\end{equation}

The NN is trained to approximate PCE coefficients while only using the limited number of measurements, $T\in \mathds R$, as target data. The mean-squared error (MSE) loss function for episode, $e$, and batch size, $B$, is defined as:
\begin{equation}
\begin{aligned}
  L_{e}(t_b, z_b) &= \frac{1}{B}\sum_{b=1}^B \lvert\lvert T_b - \hat T_{\vec\alpha} \rvert\rvert_2^2, \\
   &= \frac{1}{B}\sum_{b=1}^B \lvert\lvert T(t_b, z_b; \vec\xi_b) - \sum_{j=0}^{\lvert A\rvert} \hat C_{\vec\alpha_j}(t_b, z_b)\Psi_{\vec{\alpha_j}}(\vec\xi_b) \rvert\rvert_2^2,\\
\end{aligned}
\label{eq:loss_k}
\end{equation}
where the realizations of random vectors, $\vec\xi_b\sim\mathcal N(0,1)^{n}$ are shared between the target and approximated solution. The batch size is chosen to fit one solution sample, $B=n_t n_z$ with t-grid size, $n_t$, and z-grid size, $n_z$.

\section{Results}\label{sec:results}

\begin{figure}[t]%tbp!]%t]
 \centering
 \begin{subfigure}{.4\columnwidth}
  \centering
      \includegraphics [trim=0 0 0 0, clip, width=1.\textwidth, angle = 0]{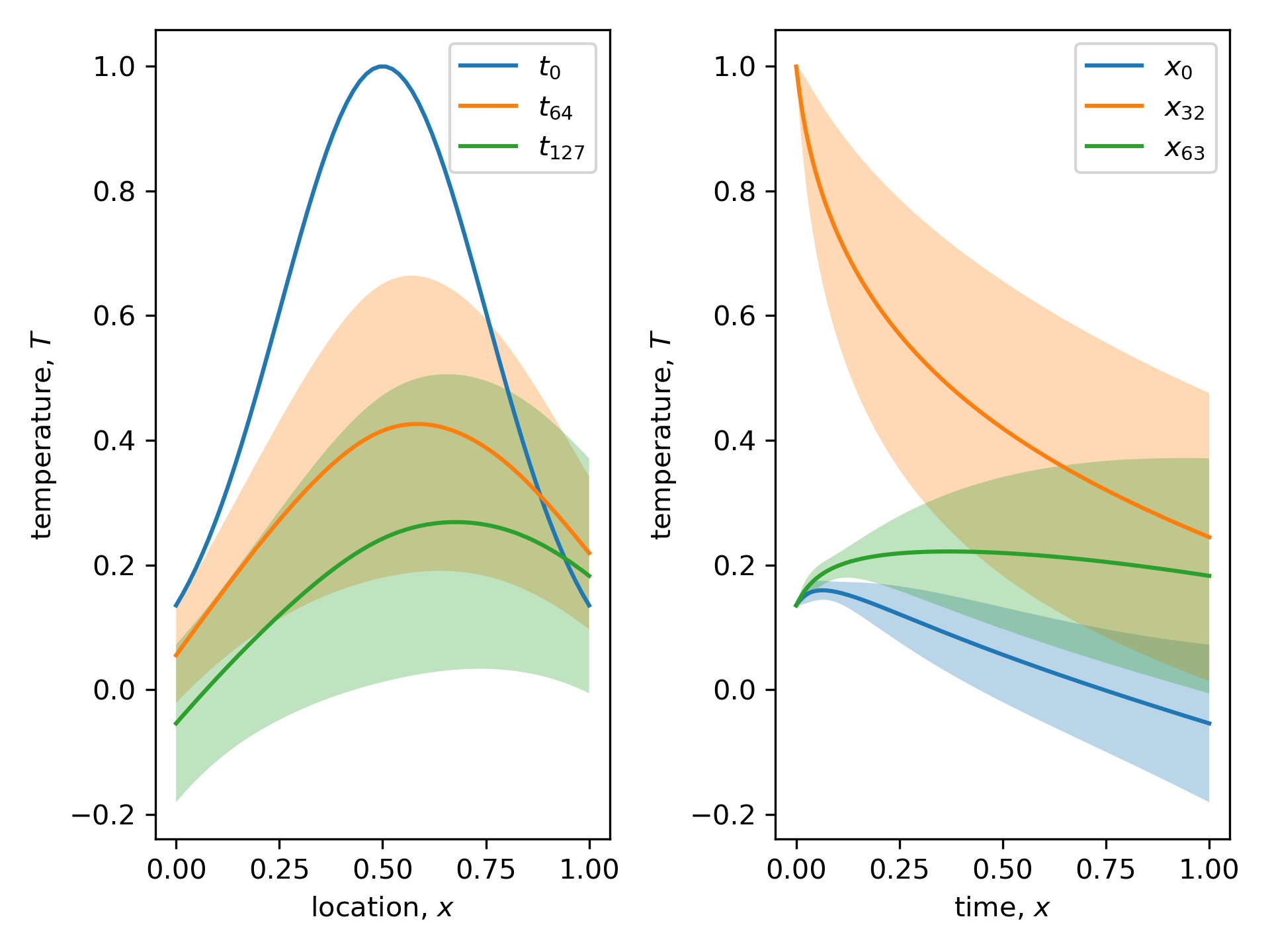}
    \caption{Target solution}
    \label{fig:target_solution}
  \end{subfigure}
\begin{subfigure}{.4\columnwidth}
    \centering
      \includegraphics [trim=0 0 0 0, clip, width=1.\textwidth, angle = 0]{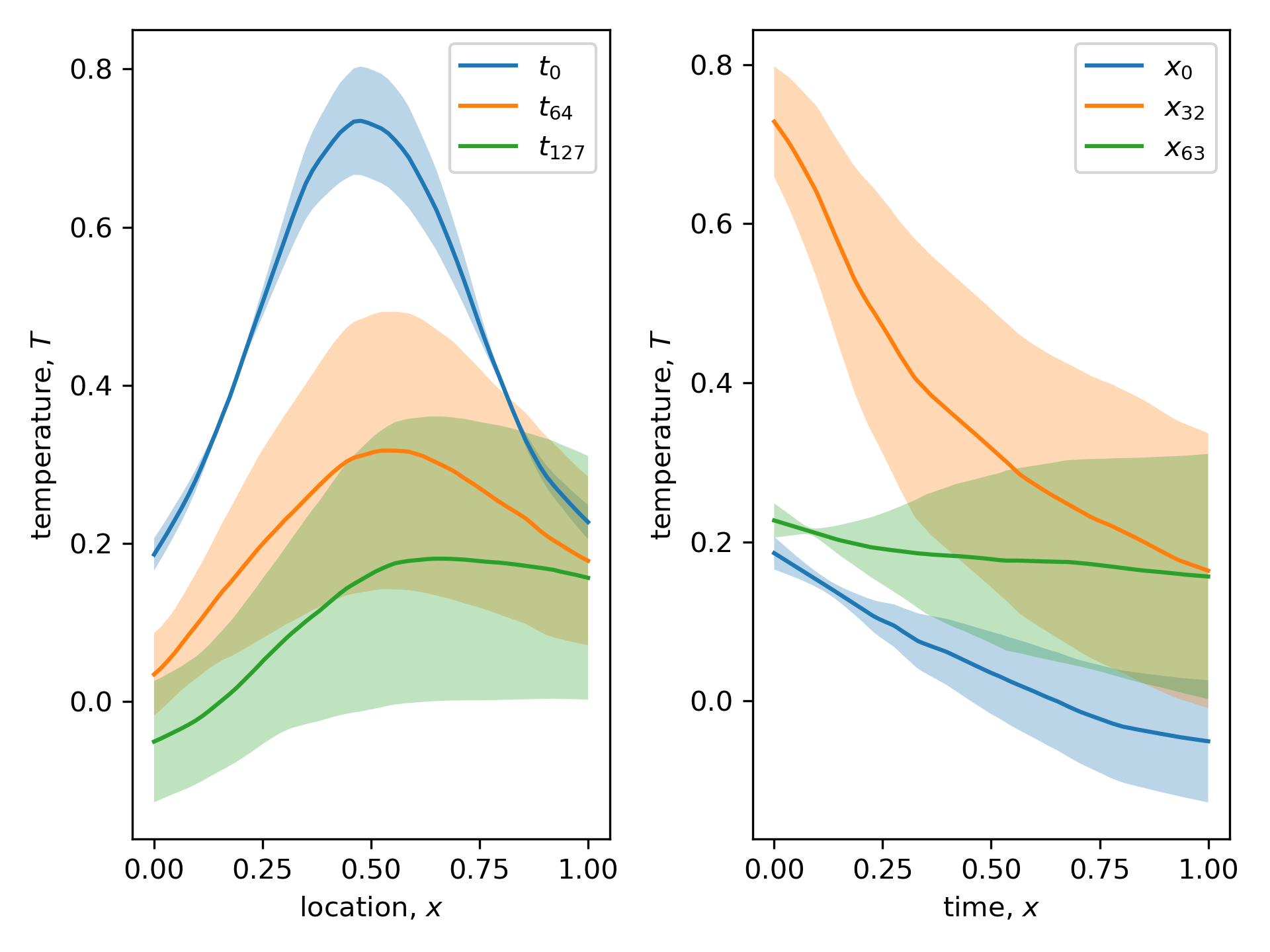}
    \caption{PCE-PINN solution (ours)}
    \label{fig:pce_pinn}
    \end{subfigure}
    \vspace{-.1in}
\caption[Solution]{\textbf{PCE-PINNs.} Initial results show that the PCE-PINNs in ~\cref{fig:pce_pinn} can approximately match the mean (line) and standard deviation (shade) of the target solution in~\cref{fig:target_solution} with $MSE=0.0078$. Importantly the approximated standard deviation also captures the growing trend towards the center location ($x=0.5$) and growing time, $t=1$.}  \label{fig:solution} 
\vspace{-.25in}
\end{figure}
\Cref{fig:solution} shows that the PCE-PINNs in~\cref{fig:pce_pinn} can successfully approximate the mean and standard deviation of the target solution~\cref{fig:target_solution}. 
Importantly the explicit formulation as polynomial chaos expansion allows us to compute the mean and standard deviation without any sampling as a function of the PCE coefficients, e.g., $\mu_T(t,z) = C_[0,0,0](t,z)$. We can note that the PCE-PINN-approximated standard deviation captures the growing trend towards the center location ($x=0.5$) and increasing time ($t=1$). Quantitative analysis shows that the mean error is, as a sum over the full spatio-temporal domain, low $MSE=0.0078$. 

We observe that the PCE-PINNs slightly overestimate the uncertainty of the initial state (blue), have a marginal positive bias towards the right boundary ($x\approx 1$), and have lower curvature during the initial steps ($t\approx 0$). Future work will explore stronger constraints on satisfying the underlying physics equations and  explore a broader choice of neural networks hyperparameters to further reduce the error.

We used a $2$-layer $128$-unit fully-connected neural network with ReLu activation. The network was trained with the ADAM optimizer with learning rate, $lr=0.001$, and $\beta=[0.9, 0.999]$ for $E=15$ epochs. The target data was generated with $n_t=128$ temporal and $n_z=64$ grid points and $n_s = 100$ samples of the solved differential equation. The maximum polynomial degree was chosen to be, $n=2$, s.t. the number of PCE coefficients, $\lvert A\rvert=3$. 

Leveraging neural network-based surrogate models can not only reduce computational complexity but also storage complexity. Our network contains $n_w = 2 n_{units}+n_{layers}n_{units}^2+n_{units}\lvert A\rvert = 33408$ weights which occupy as floats $n_{weights}4B\approx133kB$.

%\subsection{mean and std dev of solution T}

%\subsection{learned vs target PCE coefs}

\section{Discussion and future work}
We have demonstrated a novel technique for fast uncertainty propagation with physics-informed neural networks on the local advection-diffusion equation. The \name/ uses neural networks to learn the spatio-temporal coefficients of the polynomial chaos expansion, reducing the analytical and computational complexity of previous methods. Our method learned a lightweight surrogate model of the local advection-diffusion equation and successfully quantified the output uncertainties, given known parameter uncertainties. 

We note that our results show room for improvement. Future work will explore stronger constraints on satisfying the physical laws in~\cref{eq:ade}, e.g., via physics-based regularization terms~\citep{Raissi_2018} or hard physics-constraints~\citep{Beucler_2021}. Further, computational resources were limited during this experiment and future work will further optimize the choice of hyperparameters for the neural network. Lastly, the proposed approach requires computation of a training dataset of solved differential equation for a set of parameter samples which can quickly become computationally expensive. Future work, will explore self-supervised learning approaches to enable learned surrogate models without the use of  expensive training data. 

%!!! cannot feed in kappa; only propagate the distribution of kappa. Need to re-learn for different distribution of kappa...!!!

%no training data

\subsubsection*{Acknowledgments}
The authors greatly appreciate the discussions with Chris Hill, Hannah Munguia-Flores, Brandon Leshchinskiy, Nicholas Mehrle, Yanni Yuval, Paul O'Gorman, and Youssef Marzouk. 

Research was sponsored by the United States Air Force Research Laboratory and the United States Air Force Artificial Intelligence Accelerator and was accomplished under Cooperative Agreement Number FA8750-19-2-1000. The views and conclusions contained in this document are those of the authors and should not be interpreted as representing the official policies, either expressed or implied, of the United States Air Force or the U.S. Government. The U.S. Government is authorized to reproduce and distribute reprints for Government purposes notwithstanding any copyright notation herein.
\bibliography{biblio}
\bibliographystyle{iclr2021_conference}

\appendix
%\section{Appendix}
%You may include other additional sections here.

\end{document}